\pdfoutput=1
\documentclass[sigconf,nonacm]{acmart}
\makeatletter
\global\@ACM@balancefalse
\makeatother
\usepackage{tabularx}
\usepackage{needspace}

\setcopyright{none}
\renewcommand\footnotetextcopyrightpermission[1]{}
\makeatletter
\def\@mkbibcitation{%
  \par\vskip\medskipamount\noindent
  \parbox[t][84pt][t]{\columnwidth}{%
    \small
    \hrule height 0.45pt\relax
    \vskip 5pt\relax
    {\bfseries Research Status\par\nobreak}%
    \noindent This full paper has been accepted for publication in the
    proceedings of \href{https://geoindustry.github.io/geoindustryworkshop2026/}{GeoIndustry~2026},
    the 5th ACM SIGSPATIAL International Workshop on Spatial Big Data and AI
    for Industrial Applications. It will be presented in person on
    November~3, 2026, 11:30--11:45~a.m., in Riverside, California
    (Session~3: Agentic \& Explainable GeoAI). The paper will be available
    in the ACM Digital Library once the workshop proceedings are published.\par
  }%
  \par
}
\makeatother

\AtBeginDocument{%
  }
\makeatletter
\def\@subtitlefont{\footnotesize\itshape}
\makeatother
\makeatletter
\def\@typeset@author@bx{\bgroup\hsize=\author@bx@wd
  \def\and{\par}\normalbaselines
  \global\setbox\author@bx=\vtop{\if@ACM@sigchiamode\else\centering\fi
    \@authorfont\@currentauthors\par\@affiliationfont
    \@currentaffiliation\par\vskip 11pt}\egroup
  \box\author@bx\hspace{\author@bx@sep}%
  \gdef\@currentauthors{}%
  \gdef\@currentaffiliation{}}
\makeatother

\begin{document}
\settopmatter{printacmref=true}

\title{Agentic AI Orchestration Framework for Spatial Intelligence: Supply Chain Deployment}
\author{Constantinos Papantoniou}
\authornote{Corresponding author: Constantinos Papantoniou $\cdot$ \href{mailto:contact@anassa.ai}{contact@anassa.ai}. ORCID: \href{https://orcid.org/0000-0003-2115-2567}{0000-0003-2115-2567}. LinkedIn: \url{https://www.linkedin.com/in/constantinos-papantoniou/}}
\affiliation{%
  \institution{Anassa.ai}
  \city{}
  \country{USA}}
\author{Brian Hilton}
\authornote{ORCID: \href{https://orcid.org/0009-0005-1836-9927}{0009-0005-1836-9927}. LinkedIn: \url{https://www.linkedin.com/in/brian-hilton-ph-d/}}
\affiliation{%
  \institution{Research Scientist}
  \city{}
  \country{USA}}
\renewcommand{\shortauthors}{Papantoniou and Hilton}
\begin{abstract}
Industrial spatial decisions are assembled across systems that share neither a reference frame nor an owner, and analysts spend substantial effort reconciling those systems rather than reasoning about space. This paper presents ANASSA, a reference architecture integrating structured reasoning, multi-agent workflow orchestration, authoritative spatial grounding, and feedback-driven execution, and reports its instantiation in an operating supply-chain deployment. The architecture places typed data contracts at agent boundaries, validates spatial claims against a certified authoritative foundation, and enforces two blocking human-in-the-loop approvals inside a six-step cognitive loop. The deployment covers multi-hazard risk assessment on an ocean-and-inland freight lane, including cold-chain excursion detection over container telemetry. Across a three-month proof of concept (POC) with three implementation iterations, three deployment outcomes are reported: 25\% improvement in data quality, 50\% reduction in spatial workflow runtime, and 60\% lower decision latency relative to the preceding manual workflow. Absolute deployment-scale quantities and baseline counts are confidential; reported percentages are observed relative changes. These are uncontrolled deployment observations, not controlled experiments. We therefore specify instrumentation and ablation protocols for mechanism-level evaluation.
\end{abstract}
\maketitle
\newpage 
\enlargethispage{-2\baselineskip} 

\section{Introduction}
Modern geospatial operations face persistent coordination challenges across organizational boundaries. Spatial data is distributed through disconnected GIS platforms, satellite feeds, IoT sensors, field devices, cloud databases, and legacy infrastructure, each operating largely in isolation \cite{papantoniou2021enterprise}. Human analysts must extract, transform, route, and reconcile data across fragmented systems, diverting effort from spatial reasoning and decision support \cite{kaza2026, mansourian2026}. Agentic AI increasingly allows coordination tasks to be delegated to systems capable of planning, memory, and tool orchestration \cite{wang2024,xi2023}.

Recent work combines multi-agent architectures with retrieval-augmented generation to address the inability of text similarity alone to represent topology, proximity, and spatial dependency \cite{hu2025,liang2026,mansourian2026}. Other systems constrain reasoning with explicit workflow graphs encoding spatial entities and transformation dependencies \cite{bao2026,bhattaram2025}. Execution benchmarks show that agents can handle moderately complex workflows but degrade on multi-step tasks requiring precise parameter inference and tool sequencing \cite{hashemi2026,krechetova2025,wu2025,yu2026}.

This paper introduces ANASSA (Autonomous Neural Agents for Spatial Systems Architecture), an autonomous orchestration layer for geospatial intelligence, and reports its instantiation in an operating supply-chain deployment. Building on prior research \cite{papantoniou2021enterprise}, ANASSA extends the GeoBlockchain coordination model \cite{papantoniou2021geoblockchain} from blockchain-based to AI-agent-based orchestration, distributing reasoning, planning, execution, and validation across autonomous agents under human-in-the-loop governance. The architecture addresses the limitations above by integrating spatial reasoning with execution-aware orchestration, embedding feedback-driven adaptation through a continuous cognitive loop, and incorporating governance for hallucination management.

This paper makes three contributions. First, it specifies ANASSA as a unified architecture of eleven components across four layers, in which every boundary between agents is a typed data contract rather than an ad hoc handoff. Second, it instantiates that architecture in an operating supply-chain deployment---multi-hazard risk assessment on an ocean-and-inland freight lane, including cold-chain excursion detection over container telemetry---and reports three deployment outcomes together with an explicit account of what those observations do and do not support. Third, it specifies the instrument set and ablation protocol required to convert uncontrolled deployment observations into attributable, controlled evidence, and identifies design principles for agentic GIS that follow from the deployment.

The evidentiary status of this work should be stated at the outset. The outcomes reported in Section 5 are observations from operating the platform against a pre-existing manual workflow, not controlled experiments, and they do not establish which architectural mechanism produced the improvement. Sections 5.2, 5.3, and 7 state the threats to validity, the instrument set, and the ablations that would.

\section{ANASSA Framework}
Agentic GIS systems improve automation, workflow generation, and natural-language interaction \cite{hu2025,liang2026,mansourian2026}, but robust spatial reasoning, execution reliability, and evaluation remain fragmented. Multi-agent architectures improve decomposition yet remain sensitive to ambiguous spatial constraints and long dependency chains \cite{liang2026,wu2025,bao2026,yu2026}; structured reasoning approaches enforce logical consistency but do not fully integrate execution feedback \cite{bao2026,bhattaram2025}.

\begin{table}[t]
\caption{Properties explicitly specified in cited systems. Y = explicitly specified; P = described but not operationally specified; N = not specified; -- = not applicable to the benchmark purpose.}
\label{tab:prior}
\scriptsize
\setlength{\tabcolsep}{2.2pt}
\begin{tabular}{lccccc}
\toprule
System & Dec. & Contracts & Ground. & Gov. & Eval.\\
\midrule
GeoAgent \cite{hu2025} & Y & N & P & N & bench.\\
GeoFlow \cite{bhattaram2025} & Y & P & N & N & bench.\\
Spatial-Agent \cite{bao2026} & Y & P & P & N & bench.\\
GeoAgentic-RAG \cite{liang2026} & Y & N & P & N & bench.\\
Mansourian \cite{mansourian2026} & Y & N & N & N & bench.\\
GeoBenchX \cite{krechetova2025} & -- & -- & reject & -- & bench.\\
GeoAgentBench \cite{yu2026} & -- & -- & verify & -- & bench.\\
ANASSA & Y & Y & Y & Y & deploy.\\
\bottomrule
\end{tabular}
\end{table}
Classifications are based on properties explicitly documented in the cited publications; the cited references provide the basis for external verification.

\begin{figure}[t]
\centering\includegraphics[width=\columnwidth]{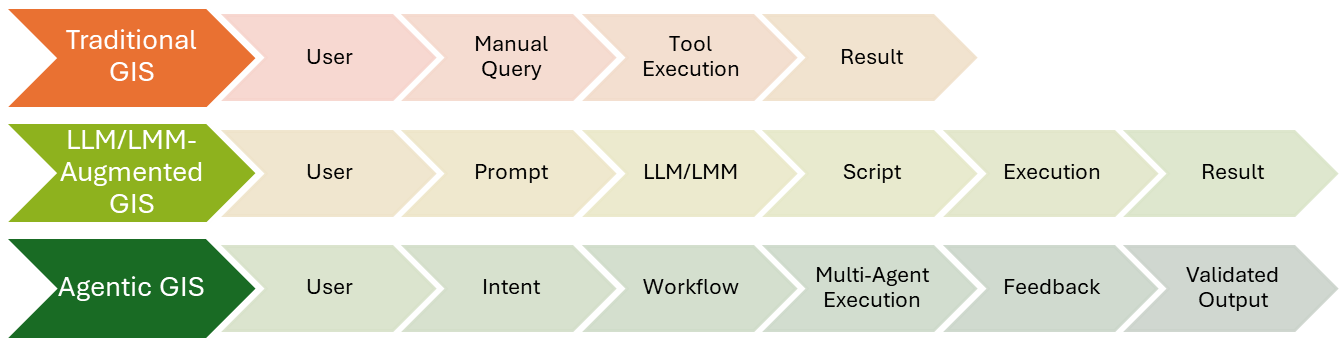}
\Description{Comparison of manual GIS, LLM-augmented GIS, and agentic GIS workflows, emphasizing the transition from analyst-coordinated steps to a closed agentic reasoning and execution loop.}
\caption{Types of GIS workflows.}
\label{fig:workflows}
\end{figure}

Figure~\ref{fig:workflows} shows the progression from manual GIS to LLM/LMM-augmented GIS and then to agentic GIS, where reasoning, multi-agent execution, feedback, and validation form a closed loop.

\subsection{Architecture Components}
\begin{figure*}[t]
\centering\includegraphics[width=\textwidth]{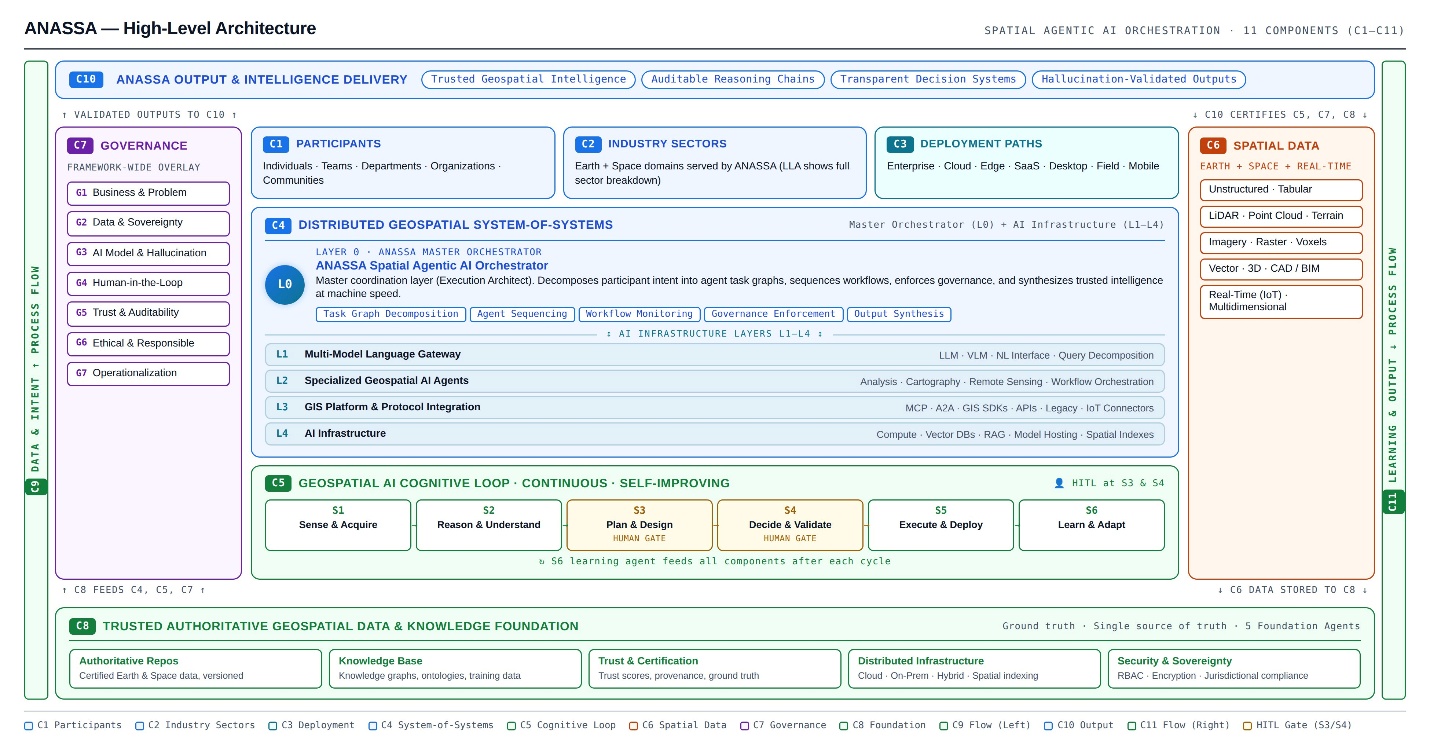}
\Description{ANASSA reference architecture showing eleven components across governance and output, operational, cognitive engine, and foundation layers, with downward and upward process flows.}
\caption{ANASSA high-level architecture.}
\label{fig:arch}
\end{figure*}

ANASSA is organized into eleven numbered architectural components (C1--C11) structured across four layers (Figure~\ref{fig:arch}). The Governance and Output layer (C7, C10) validates framework outputs and governs all agent and workflow behavior. The Operational layer (C1--C4, C6) defines participants, industry sectors, deployment paths, the technical system-of-systems backbone, and multimodal spatial data. The Cognitive Engine (C5) runs the six-step Geospatial AI Cognitive Loop. The Foundation layer (C8) provides trusted authoritative data and knowledge that all other components draw from, while C9 and C11 represent the continuous downward and upward process flows that close the loop. The layer assignment is summarized in Table~\ref{tab:layers}.

\begin{table}[t]
\caption{Architectural layers and components.}\label{tab:layers}\footnotesize
\begin{tabularx}{\columnwidth}{l l X}\toprule
Layer & Components & Function\\\midrule
Governance/Output & C7,C10 & Validates outputs and governs agents/workflows\\
Operational & C1--C4,C6 & Participants, sectors, deployment, system-of-systems, spatial data\\
Cognitive Engine & C5 & Continuous sensing, reasoning, planning, decision, execution, learning\\
Foundation & C8,C9,C11 & Trusted data/knowledge and process-flow connections\\\bottomrule
\end{tabularx}
\end{table}

Two numbering schemes are kept separate throughout. Architectural layers are the four groupings of Table~\ref{tab:layers}, and each component belongs to exactly one. Infrastructure levels C4-L0 through C4-L4 are internal to Component 4: C4-L0 is the Orchestrator, C4-L1 the multi-model language gateway, C4-L2 the specialized geospatial agents, C4-L3 the GIS platform and protocol integration surface, and C4-L4 the AI infrastructure. The Orchestrator is therefore Level 0 of an Operational-layer component, not a fifth architectural layer. A third scheme, the agent coordination levels A0--A4, is introduced in Section 3.3. Each component is implemented as a coordinated set of specialized agents under C4-L0, which decomposes participant intent into agent tasks, sequences execution, monitors outputs, and synthesizes results into trusted intelligence. After each cycle, the Learning Agent may update authorized operational parameters and generate governance-rule proposals; it cannot autonomously weaken or bypass A3 controls. The ``Neural'' in the acronym refers to the LLM- and LMM-based reasoning agents at C4-L1 and C4-L2. ANASSA proposes no novel neural architecture and makes no claim about model internals; the contribution is the coordination, contract, and governance structure surrounding those models.

\subsection{Geospatial AI Cognitive Loop}
Component C5 is the core of ANASSA: a continuous six-step reasoning and action cycle that transforms spatial data into spatial intelligence and intelligence into action (Table~\ref{tab:loop}). The six steps are S1 Sensing, S2 Reasoning, S3 Planning, S4 Decision and Hallucination Validation, S5 Execution, and S6 Learning. All six are coordinated by the Orchestrator (C4-L0) and grounded in authoritative data from C8. The cycle is feedback-driven rather than linear. Outputs from S5 do not terminate the workflow; they are captured at S6 and fed back into the system, enabling authorized operational parameters and agent capabilities to adapt while governance-policy changes remain subject to explicit approval. This closed-loop design distinguishes ANASSA from pipeline-based agentic GIS systems, which typically treat execution as a terminal stage and rely on offline retraining for adaptation \cite{liang2026,yu2026}.

\begin{table}[t]
\caption{Geospatial AI Cognitive Loop (C5).}\label{tab:loop}\footnotesize
\begin{tabularx}{\columnwidth}{c l X l}\toprule
Step & Agent & Function & HITL\\\midrule
S1 & Sensing & Receives prepared data from C6, detects spatial events and anomalies warranting cognitive processing, and reports status to the Orchestrator & No\\
S2 & Reasoning & Applies spatial intelligence to establish context and pattern, and selects models, tools, and workflows & No\\
S3 & Planning & Generates geospatial workflows and solution strategies aligned to participant needs and industry requirements (C1, C2) & Blocking\\
S4 & Decision/validation & Evaluates options and validates outputs against authoritative ground truth, checking every spatial claim against C8 & Blocking\\
S5 & Execution & Deploys through C3 infrastructure, publishes maps and dashboards, triggers downstream workflows, and notifies stakeholders & No\\
S6 & Learning & Captures outcomes and feeds improvements to all components, updating Orchestrator coordination rules & No\\\bottomrule
\end{tabularx}
\end{table}

Two mandatory human-in-the-loop gates appear at S3 (Planning) and S4 (Decision and Hallucination Validation). At S3, participants review and approve the generated workflow before any geospatial operation is executed, preventing inappropriate model or tool selection from propagating downstream. At S4, participants approve or reject the cross-validated outputs before publication. Both gates are blocking rather than advisory in the deployment reported here, and both record approver identity and timestamp. These gates reflect the framework's position that the highest-stakes geospatial decisions must remain under explicit human accountability. Section 6 examines the cost this imposes on machine-pace orchestration, which is the subject of RQ4.

\subsection{Governance and Hallucination Management}
Component C7 functions as the governance backbone of ANASSA and is positioned as a framework-wide overlay connecting to every other component. Its design is informed by the CPMAI methodology \cite{cognilytica2019}, which structures the AI project lifecycle from problem definition through production operation. Governance is organized across seven pillars: business and problem governance (G1), data governance and sovereignty (G2), AI model and agent governance with hallucination management (G3), human-in-the-loop governance (G4), trust and auditability (G5), ethical AI and responsible use (G6), and operationalization and continuous improvement (G7). A distinctive feature of the framework is explicit management of geospatial hallucinations as a primary governance concern. Fabricated coordinates, invented boundaries, and false spatial claims can produce incorrect maps used for high-stakes decisions, generate jurisdictional disputes, and in safety-critical contexts endanger lives. The Claim Validation Agent in C7 tests every spatial claim in a candidate output against the certified ground truth maintained in C8. Claims that do not match a certified assertion are not silently dropped and not silently published: they are delivered with the output on an explicit unverified-claim list. This mechanism is specified as contract K5 and is the subject of RQ2.

\section{Research Questions and Use Case}
The central research question motivating this work is: how can agentic GIS systems integrate spatial reasoning, workflow orchestration, and execution feedback to enable robust and scalable geospatial analysis in real-world environments? This paper investigates that question through a single instantiated use case: multi-hazard supply-chain risk assessment on an operating ocean-and-inland freight lane. The use case requires no new sensing infrastructure; it takes existing telemetry, network, and hazard streams as its starting point and examines what the orchestration and spatial reasoning layer contributes on top of them. The four research questions below are scoped to what this deployment can and cannot answer, and Section 6 reports the evidence obtained against each.

\textbf{RQ1 Coordination.} Do typed contracts replacing analyst mediated handoffs reduce coordination effort, measured by analyst time and systems touched per analysis? \textbf{RQ2 Grounding.} Does validating spatial claims against certified authoritative data reduce spatial reference and unsupported claim errors and detect them earlier? \textbf{RQ3 Fusion and reproducibility.} Does versioned fusion of telemetry, network state, and asynchronously updating hazards enable auditable decisions requiring multiple sources? \textbf{RQ4 Governance under machine pace.} Can blocking HITL gates preserve accountability without eliminating the decision-latency advantage of orchestration?

\subsection{Use-Case Requirements and Failure Model}
A logistics planner evaluating supply-chain vulnerability must combine warehouse locations, transport routes, and environmental risk such as flood zones, which requires spatial joins, distance analysis, and risk overlay across multiple steps. Existing LLM-based systems can generate such workflows but fail when an intermediate output is missing or a dependency breaks, and the failure can be silent: the workflow produces a plausible but unsupported result rather than an error. Under ANASSA the Orchestrator monitors execution, detects the missing layer, and replans, because every transition is a typed contract with a declared precondition. A missing coordinate reference system, unresolved owner, expired restriction, ungrounded facility, or trajectory gap therefore becomes a contract violation at the point of handoff rather than a silent degradation downstream. The use case exercises execution reliability, adaptability, and alignment between automated reasoning and human-governed decision making. It does not exercise cross-domain generality, which Section 7 identifies as future work.

RQ4 deliberately states a tension rather than presuming that governance is free. Two mandatory approvals inside a machine-pace loop can reduce the very latency advantage the architecture seeks to create. Decision latency is consequently defined from objective to \emph{approved} decision, not merely query to generated result, so the cost of human review remains inside the measured interval.

\subsection{Supply-Chain Agent Specification}
Coordination levels A0--A4 are distinct from C4 infrastructure levels. A0 is the Orchestrator; A1 the cognitive-loop roles; A2 domain specialists; A3 governance; and A4 foundation services. In this deployment, A4 is realized by C8 services rather than a distinct agent. Table~\ref{tab:agents} summarizes the eleven active roles and representative validation conditions.

\begin{table*}[t]
\caption{Supply-chain agent roles and validation conditions.}
\label{tab:agents}\footnotesize
\begin{tabularx}{\textwidth}{c l l X X}
\toprule Level & Agent & Origin & Function & Validation condition\\\midrule
A0 & ANASSA Orchestrator & Platform & Decomposes intent into a task graph, sequences agents, and enforces governance & Every node declares an input contract; an unresolved node blocks rather than proceeds\\
A1 & Cold-Chain Telemetry (C5-S1) & Platform & Ingests reefer and container telemetry from C6 & Bundle rejected if CRS is absent; a series gap is a monitoring failure, never interpolated\\
A1 & Learning (C5-S6) & Platform & Updates preference models, parameters, and agent capabilities each cycle & Every update carries a cycle identifier and lists changed parameters\\
A2 & Administration & Org role & Maintains organizational, contractual, and jurisdictional context & A consignment must resolve to a legal owner before downstream agents act\\
A2 & Supply Origin & Org role & Resolves supplier facility geometry and dispatch readiness & Facility matched to a certified C8 address point, or the claim is unverified\\
A2 & Port State & Org role & Maintains terminal state at load and discharge ports & Terminal geometry from certified C8 assertions, not carrier free text\\
A2 & Maritime Leg & Org role & Produces ocean corridor geometry and arrival estimates & Gap-aware trajectories; no interpolation across a gap above the lane threshold\\
A2 & Inland Leg & Org role & Routes the final haul to the distribution center & Restrictions carry a validity interval; expired ones are not reused\\
A2 & Inventory and Allocation & Org role & Reconciles arriving quantity and proposes reallocation & Every proposal cites the observation identifiers that triggered it\\
A3 & Claim Validation & Platform & Tests each spatial claim against certified C8 assertions & Each claim is matched to a certified assertion or listed as unverified\\
A3 & HITL Gate & Platform & Enforces the blocking approvals at C5-S3 and C5-S4 & No publication without approver identity and timestamp; rejection returns to S3 with a reason code\\
\bottomrule
\end{tabularx}
\end{table*}

\subsection{Coordination Semantics}
The coordination levels define invocation authority as well as role grouping. A0 decomposes intent and sequences work; A1 performs the cognitive-loop functions; A2 supplies domain-specialist execution; A3 governs publication and can block; A4 supplies authoritative foundation services. Governance agents are intentionally asymmetric with execution agents: A3 may stop or return work to the cognitive loop, while A1 and A2 cannot bypass or disable an A3 decision. In the reported instantiation, A4 functions are provided by C8 services rather than a separately autonomous agent. K8 may update only pre-authorized operational parameters; changes to blocking gates, validation requirements, publication authority, or governance policy require explicit authorized approval.

The validation conditions in Table~\ref{tab:agents} are part of the architecture rather than implementation commentary. Each condition names a checkable failure mode: legal ownership must resolve before downstream action, facilities and terminals must resolve to certified geometry or be marked unverified, trajectories must remain gap-aware, restrictions must carry validity intervals, and reallocation proposals must cite the observations that triggered them. These conditions externalize a portion of expert operating knowledge into machine-checkable boundaries while leaving consequential judgment at the HITL gates.

\section{Implementation}
\subsection{Deployment Architecture}
The implementation is a hybrid multi-cloud deployment integrating foundation-model reasoning, deterministic spatial services, automated workflows, and operator dashboards. Provider-specific model identifiers and infrastructure configuration are deployment-specific and are not reported; the contribution evaluated here is the orchestration and governance layer rather than model-specific performance. Container telemetry is written to versioned storage through C6$\rightarrow$C8. Certification denotes workflow-authorized evidence with known source identity, spatial reference, temporal validity where applicable, and recorded storage version; it does not imply that a source is error-free. Spatial/vector indexes are version-recorded and rebuilt rather than served stale.

The workload is spatial big data rather than merely workflow automation. Container telemetry is a high-cardinality point stream keyed by consignment; active-consignment counts, observation rates, retention windows, and authoritative-layer version counts are confidential deployment-scale quantities. Tracks require gap-aware trajectory handling, while hazard polygons update asynchronously and carry validity intervals. The resulting join is spatio-temporal and runs against version-recorded spatial indexes. Typed contracts declare this join once rather than forcing analysts to reconstruct it per query.

\subsection{Typed Contracts}
Every agent boundary is a typed data contract with a declared precondition. Table~\ref{tab:contracts} specifies the eight contracts and the failure modes they prevent.
\begin{table*}[t]
\caption{Supply-chain data contracts.}
\label{tab:contracts}\footnotesize
\begin{tabularx}{\textwidth}{c l X X X}
\toprule & Boundary & Payload & Precondition & Failure mode prevented\\\midrule
K1 & C6$\rightarrow$C8 & Telemetry and hazard writes to the foundation & Certification level assigned at write; storage version recorded & Uncertified data indistinguishable from certified\\
K2 & C6$\rightarrow$C5-S1 & Telemetry bundle for sensing & Explicit CRS on every bundle; gaps flagged as monitoring failures & A missing window smoothed into a fictitious history\\
K3 & C5-S2$\rightarrow$C5-S3 & Candidate workflow as a task graph & Each node declares tool, parameters, and expected output type & Inapplicable tool selection propagating undetected\\
K4 & HITL gate at S3 & Plan approval & Blocking; approver identity and timestamp recorded & Execution of a plan no accountable human reviewed\\
K5 & C8$\rightarrow$C5-S4 & Grounding assertion set & Each claim matched to a certified assertion or listed as unverified & Fabricated coordinates published as analysis\\
K6 & HITL gate at S4 & Decision approval & Blocking; rejection returns to S3 with a reason code & Unaccountable publication of a high-stakes output\\
K7 & C5-S5$\rightarrow$C10 & Validated output with provenance & Index version and query recorded & Irreproducible audit; a result that cannot be regenerated\\
K8 & C5-S6$\rightarrow$C1--C7 & Learning update & Cycle identifier attached; changed parameters enumerated & Silent drift in agent behavior between cycles\\
\bottomrule
\end{tabularx}
\end{table*}

K2 and K5 change when failures surface. K3 machine checks verify required fields, registered input/output types, dependencies, and CRS/temporal preconditions; S3 retains judgment over whether the method is operationally appropriate. K5 evaluates structured claims containing entity/geometry, CRS, valid time, predicate, evidence identifiers, source versions, and validation status. Source-level claims use identity and geometry agreement, while derived claims additionally require deterministic spatial and temporal predicates (for example, intersection with a hazard during overlapping validity). Unmatched claims remain explicitly unverified. Representative fault cases include a K2 bundle without a CRS and an expired Inland Leg restriction (Figure~\ref{fig:inland}).

\subsection{End-to-End Execution Trace}
\begin{figure*}[!t]
\centering
\begin{minipage}[t]{0.48\textwidth}
\centering
\includegraphics[height=0.35\textheight,keepaspectratio]{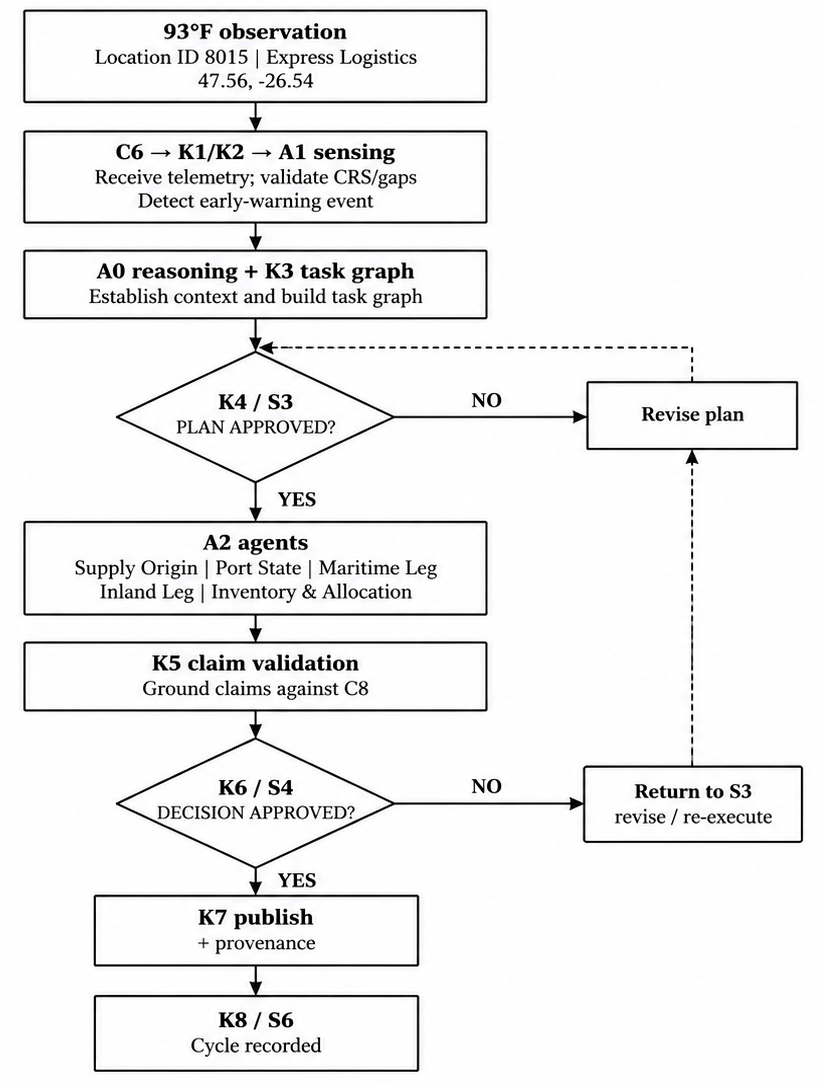}
\Description{Anonymized end-to-end trace for Location ID 8015 showing the 93 degree Fahrenheit observation, sensing and orchestration, planning approval, domain execution, claim grounding, decision approval, publication, provenance, cycle recording, and reject-revise paths.}
\captionof{figure}{End-to-end ANASSA agent and human-in-the-loop trace for the selected cold-chain observation.}
\label{fig:trace}
\end{minipage}\hfill
\begin{minipage}[t]{0.48\textwidth}
\centering
\includegraphics[height=0.35\textheight,keepaspectratio]{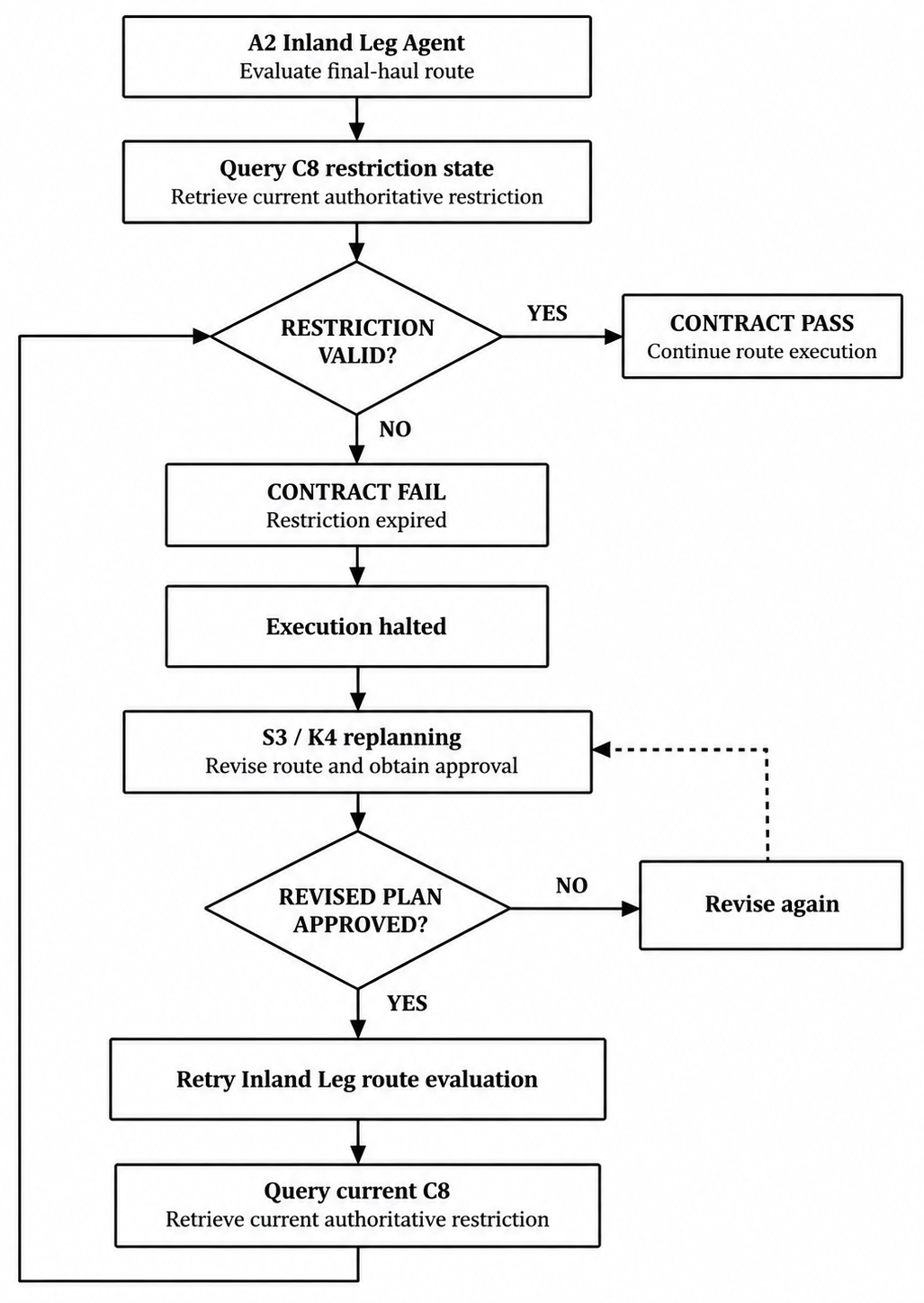}
\Description{Representative Inland Leg contract failure-and-recovery trace showing authoritative restriction retrieval, contract pass or failure, execution halt, replanning, re-approval, retry, and re-query.}
\captionof{figure}{Representative Inland Leg contract failure-and-recovery trace.}
\label{fig:inland}
\end{minipage}
\end{figure*}

The supply-chain instantiation can be read as a concrete execution trace through the contracts and cognitive loop rather than as a collection of independent agents. Figure~\ref{fig:trace} anchors the sequence to the selected cold-chain observation shown later in the operator view: Location ID 8015, Express Logistics, 93$^{\circ}$F, at 47.56, -26.54. A cycle begins when C6 receives telemetry or a hazard-layer update. K1 assigns certification state and records the storage version before the new observation enters C8. K2 then packages the sensing bundle with an explicit coordinate reference system and with gaps represented as monitoring failures rather than silently interpolated history. S1 can therefore distinguish an observed event from a missing observation window before the reasoning stage is invoked.

At S2, the Reasoning function establishes spatial and operational context and identifies the tools and domain roles needed for the objective. The Orchestrator converts that reasoning into a task graph under K3. Each node declares its tool, parameters, expected output type, and dependency. The first blocking gate, K4 at S3, exposes the candidate plan to an accountable participant before execution; rejection returns the workflow for revision before geospatial operations are committed.

Approval identity/timestamp and tool-output metadata are retained in operational logs but omitted from the anonymized trace.

Execution then proceeds through the A2 domain roles. The Supply Origin Agent resolves the dispatch facility against certified C8 geometry. The Port State Agent maintains the load and discharge terminal states. The Maritime Leg Agent produces a gap-aware ocean corridor, and the Inland Leg Agent applies restrictions with validity intervals to the final-haul route. The Inventory and Allocation Agent reconciles arriving quantity and, when required, proposes reallocation while citing the triggering observations. Each transition fires a declared contract, so a violated precondition is visible at the boundary where it occurs.

Before a consequential output can be published, K5 provides the grounding assertion set to S4. The Claim Validation Agent attempts to match each spatial claim to certified C8 evidence; unmatched claims are surfaced explicitly as unverified rather than being silently promoted to facts. K6 then applies the second blocking approval. Rejection returns the workflow to planning with a reason code, preserving both accountability and a machine-readable path for revision. Approved outputs pass through K7 with the query and spatial-index version attached to the provenance record.

Finally, S6 closes the loop through K8. Learning updates carry a cycle identifier and enumerate changed parameters, preventing silent behavioral drift between runs. Together, these stages share one auditable state-transition model.

\subsection{Multi-Agent Coordination of a Consignment}
Figure~\ref{fig:route} shows an operator view of one consignment from Orlando through Miami and Rotterdam to Venlo. Supply Origin resolves the supplier facility, Port State maintains terminal state, Maritime Leg carries the ocean corridor, Inland Leg handles final haul, and Inventory/Allocation reconciles arrival. Figure~\ref{fig:inland} expands the Inland Leg role with a representative failure-and-recovery path: an expired restriction blocks execution and returns the workflow to S3 for replanning; after re-approval, execution resumes on the updated route.

Each transition is a contract firing rather than a handoff between screens: the corridor geometry produced by the Maritime Leg Agent is the same versioned artifact later tested by the Claim Validation Agent against certified terminal geometry from C8 under K5.

\subsection{Cold-Chain Excursion Detection}
\begin{figure*}[!t]
\centering
\includegraphics[width=\textwidth]{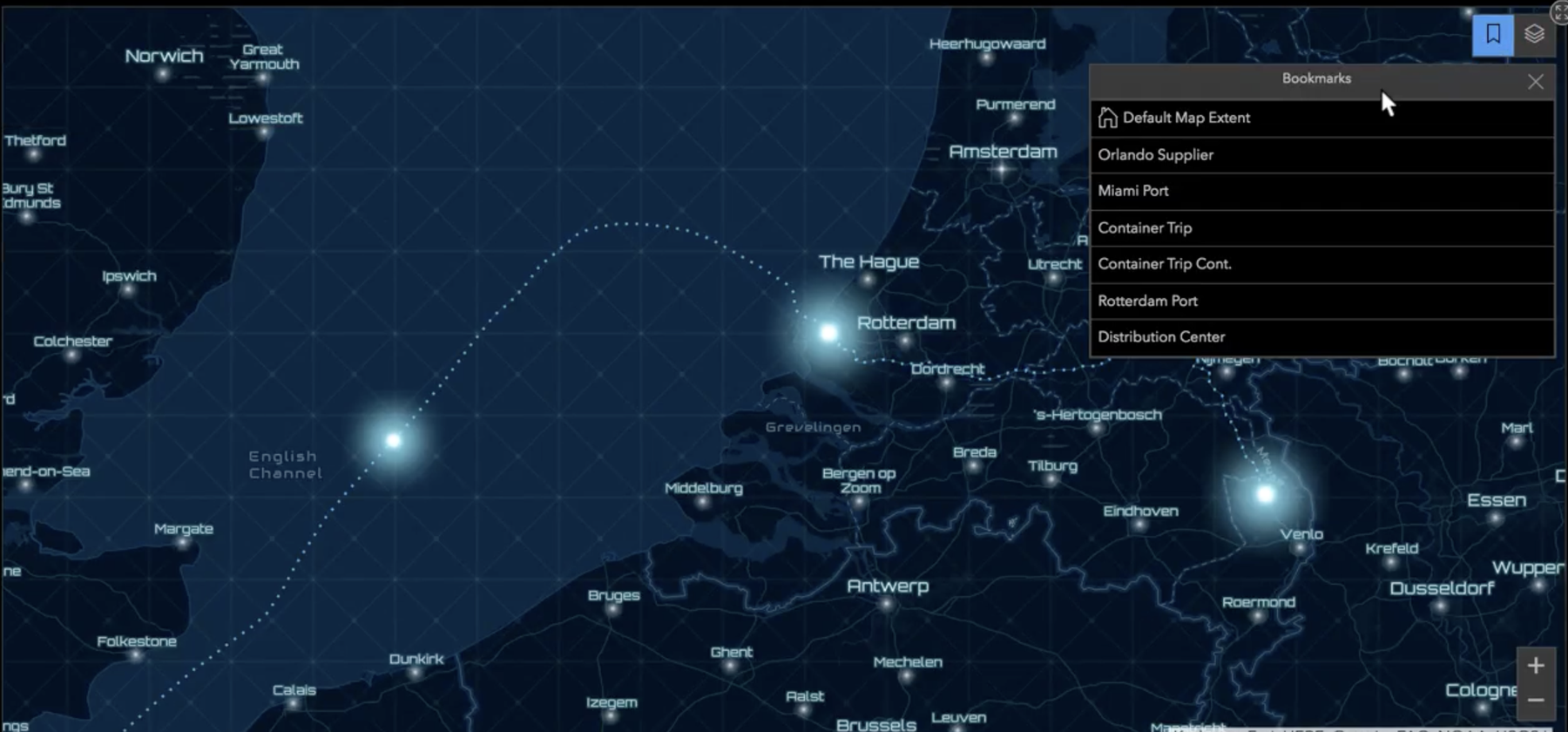}
\Description{Supply-chain consignment workflow showing spatial and operational data sources, agent processing stages, human approval gates, and resulting decision artifacts.}
\captionof{figure}{Full-width operator view of multi-agent coordination for one consignment, Orlando to Venlo. The spatial waypoints correspond to the supply-origin, port-state, maritime-leg, inland-leg, and inventory/allocation roles.}
\label{fig:route}
\par\vskip 1em\relax
\includegraphics[width=\textwidth]{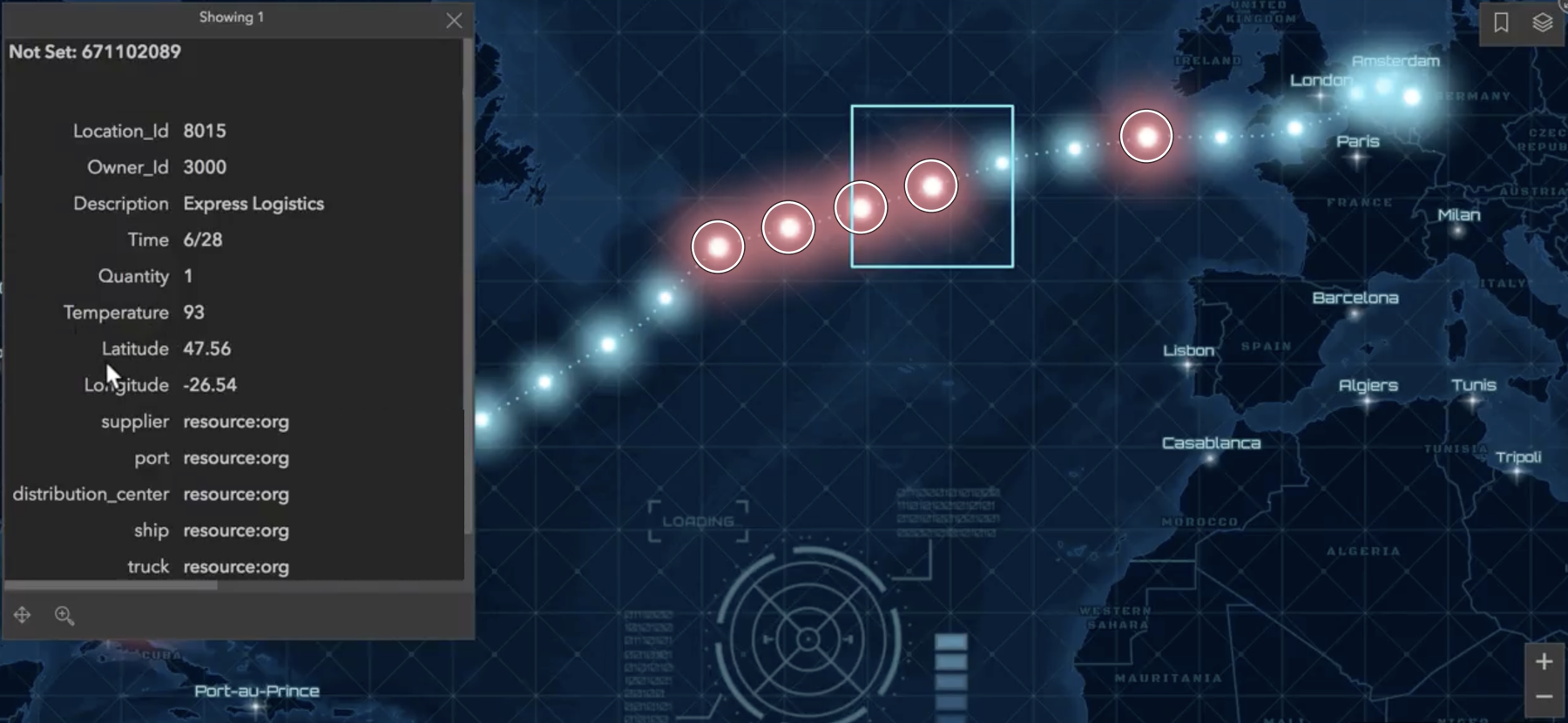}
\Description{Cold-chain workflow combining container telemetry, trajectory information, hazard context, claim validation, and governed escalation for temperature-excursion decisions.}
\captionof{figure}{Full-width cold-chain excursion view on the ocean leg. Highlighted readings define a spatially extended early-warning event rather than an isolated point alarm.}
\label{fig:cold}
\end{figure*}

A second view of the same lane (Figure~\ref{fig:cold}) surfaces a temperature excursion on the ocean leg, detected by the Cold-Chain Telemetry Agent. Temperature is reported in degrees Fahrenheit. For the illustrated canned-food POC, 95$^{\circ}$F was configured as the maximum storage threshold and the selected 93$^{\circ}$F observation was surfaced as an early-warning condition before that maximum was reached. Sampling cadence is source-determined rather than fixed by ANASSA; the framework reasons over received observations and treats missing windows as monitoring gaps rather than interpolating them. Event extent is derived from the valid trajectory segment associated with successive excursion observations, not from an inferred path across missing measurements \cite{mercier2017,ndraha2018}.

\Needspace{160pt}
\subsection{Versioning, Provenance, and Reproducibility}
The authoritative foundation is not treated as a single current-state database. Telemetry writes receive a certification level and storage version at ingestion, while hazard layers carry independent validity intervals. Spatial and vector indexes are version-recorded and rebuilt rather than silently serving stale state. Contract K7 preserves the index version and query with the validated output. Consequently, the artifact used by a domain agent and the artifact later examined by Claim Validation can be tied to the same recorded state.

This design addresses an audit problem specific to spatial decision systems: the same nominal query can return a different answer after a geometry, restriction, or hazard layer changes. Here, audit re-issue means recovering the validated decision artifact together with the recorded query, spatial-index/source versions, transformation context, and provenance chain. It does not imply bit-for-bit regeneration of the complete agentic reasoning process, which would require additional model and execution-state preservation. Quantitative re-query determinism was not recorded during the observation period; RQ3 therefore remains illustrated rather than experimentally supported.

\section{Outcomes and Industrial Impact}
\subsection{Observed Outcomes}
Three outcomes were recorded during a three-month POC that progressed through three implementation iterations. The iterations reflect successive operational refinements rather than independent experimental runs. Individual workflow-run counts, rejected/deferred-output counts, and workload distributions were not retained for controlled statistical analysis; absolute deployment quantities remain confidential. D1--D3 are therefore reported as normalized relative changes against the preceding manual workflow, not as inferential estimates. For each reported reduction, $\Delta M=(M_{\mathrm{manual}}-M_{\mathrm{POC}})/M_{\mathrm{manual}}\times100\%$.

\textbf{D1. Data quality.} Human reviewers independently reviewed outputs. Data quality is the unsupported-claim rate: unsupported published claims divided by total published claims; claims surfaced as unverified and routed to review are tracked separately. The POC was associated with a 25\% lower unsupported-claim rate against the manual workflow.

\textbf{D2. Workflow runtime.} Runtime spans initiation of the multi-hazard analytical workflow through production of the validated analytical output. The POC combined environmental hazard layers, transport-network restrictions, and cold-chain telemetry. It was associated with a 50\% reduction in workflow runtime relative to the preceding manual workflow.

\textbf{D3. Decision latency.} The same deployment was associated with 60\% lower decision latency, measured from objective to approved decision. The two blocking approvals are inside this interval; D3 therefore captures the human-governed end-to-end decision path rather than only automated execution.

Absolute baseline and post-deployment values are confidential. Table~\ref{tab:outcomes} therefore reports relative changes and measurement bases only; no undisclosed values are replaced with synthetic placeholders.
\begin{table}[t]
\caption{Deployment outcomes against the manual baseline.}
\label{tab:outcomes}\footnotesize
\begin{tabularx}{\columnwidth}{c l X l}
\toprule ID & Outcome & Measurement basis & Change\\\midrule
D1 & Data quality & Published claims lacking supporting references; human-reviewed & 25\% lower rate\\
D2 & Runtime & Wall-clock time per multi-hazard assessment & 50\% lower\\
D3 & Decision latency & Objective to approved decision; gates included & 60\% lower\\
\bottomrule
\end{tabularx}
\end{table}

\subsection{Threats to Validity}
D1--D3 are deployment observations from operating the platform against the pre-existing manual workflow as baseline. They are not controlled experiments, and several limits apply. The compared configurations differ in more than one mechanism at a time: ANASSA introduces typed contracts, authoritative grounding, and governance gates simultaneously. The observations therefore support an association between the coordination layer as a whole and material operational improvement, but not attribution to any single mechanism. The three-month POC was not randomized, progressed through three implementation iterations, and experienced varying lane conditions; seasonal, volume, and iteration effects cannot be separated from the observed changes. The deployment covers one operating context, and behavior under different data-certification regimes, jurisdictional constraints, or organizational structures is unobserved. Instrumentation was incomplete during the observation period; in particular, intervention rate and gate dwell time were not captured, which leaves the evidence against RQ4 partial. Finally, the baseline is the organization's pre-existing manual workflow, not a competing agentic-GIS system. The comparison therefore speaks to the value of orchestration over manual coordination and not to ANASSA's standing relative to GeoAgentic-RAG, GeoAgent, or other published agentic-GIS pipelines.

\subsection{Instrumentation for Controlled Evaluation}

\begin{table*}[t]
\caption{Instrumentation mapped to research questions.}\label{tab:instruments}\footnotesize
\begin{tabularx}{\textwidth}{c l X l}\toprule
ID & Instrument & Definition & RQ\\\midrule
I1 & Analyst time & Analyst-attributable wall-clock time, objective to approved decision & RQ1\\
I2 & Systems touched & Distinct systems analyst accesses directly & RQ1\\
I3 & Contract violations & Failed preconditions per thousand contract firings & RQ1,RQ2\\
I4 & Spatial-reference errors & Outputs with absent/incorrect CRS or datum & RQ2\\
I5 & Unsupported claims & Published claims lacking matching certified assertion & RQ2\\
I6 & Unverified-claim list & Claims routed to human review per analysis & RQ2,RQ4\\
I7 & Detection stage & First loop step (S1--S5) where error is caught & RQ2\\
I8 & Fusion coverage & Multi-source decisions served at accepted precision floor & RQ3\\
I9 & Re-query determinism & Audit re-queries identical to original result & RQ3\\
I10 & Audit completeness & Outputs with full provenance chain to certified sources & RQ3,RQ4\\
I11 & Intervention rate & Cycles in which a gate rejects/modifies candidate & RQ4\\
I12 & Gate dwell time & Elapsed time from gate presentation to action & RQ4\\
I13 & Decision latency & Objective to approved decision, gate cost included & RQ4\\\bottomrule
\end{tabularx}
\end{table*}

The architecture emits contract, gate, and provenance logs from which the next evaluation can compute: analyst time and systems touched (RQ1); contract violations, spatial reference errors, unsupported claims, unverified claims, and failure detection stage measures (RQ1/RQ2); fusion coverage, re-query determinism, and audit completeness (RQ3); and intervention rate, gate dwell time, and objective-to-approved-decision latency (RQ4). Unverified-claim list length is especially important: if authoritative coverage is too low, the architecture merely relocates coordination burden to human review.

\subsection{Industrial Impact}
The architecture is aimed at organizations that already hold the data and already have the problem at scale, and whose analysts spend a large share of their time reconciling systems rather than reasoning about space. The same contract set applies wherever a spatial decision is assembled across systems that share neither a reference frame nor an owner: last-mile delivery planning, where route feasibility depends on restriction attributes held by a third party; dynamic routing under disruption, where hazard layers and network state update on different cadences; and resource allocation across facilities, which is only as good as the derived locations it is computed over. What transfers is the typed boundary rather than the domain model, and coverage at an operator-accepted precision floor rather than raw throughput. Two conditions govern whether that transfer pays. The first is foundation coverage: below the threshold at which the unverified-claim list stays reviewable at operational volume, the architecture relocates coordination burden rather than reducing it. The second is that typed boundaries must be declared once by someone who understands the domain, an up-front cost paid against recurring analyst time rather than a capability the platform supplies. This is a claim about applicability, not a demonstrated result; the architecture has been instantiated in one domain only.

\section{Discussion}
The deployment partially supports RQ1 through the observed runtime and decision-latency reductions, but analyst time and systems touched were not separately recorded. RQ2 is partially supported by the composite data-quality improvement, but the grounding contribution is not isolated. RQ3 is illustrated through versioned artifacts and audit re-issue by construction, but quantitative fusion and determinism measures were not recorded. RQ4 is partially supported because blocking gates did not eliminate the observed latency advantage, but intervention rate and gate dwell time were not captured.

\begin{table*}[t]
\caption{Evidence obtained against each research question.}\label{tab:evidence}\footnotesize
\begin{tabularx}{\textwidth}{l X X X}\toprule
RQ & Mechanism under test & Evidence in this paper & Status\\\midrule
RQ1 Coordination & Typed contracts K1--K3; Orchestrator task-graph sequencing & D2 (50\% workflow runtime reduction) and D3 (60\% lower decision latency) against the manual baseline & Partially supported. Uncontrolled and confounded; I1 and I2 were not separately recorded\\
RQ2 Grounding & K2 CRS and gap enforcement; K5 claim validation; Claim Validation Agent & D1 (25\% data-quality improvement) attributed to interception of low-confidence claims & Partially supported. The composite quality measure is not decomposed into I4, I5, and I7, so the earlier-detection claim is argued rather than measured\\
RQ3 Fusion and reproducibility & C8 foundation; K7 versioned provenance; spatio-temporal join behind typed contracts & Section 4 walkthrough; version-recorded indexes support audit re-issue by construction & Illustrated only. No quantitative fusion or determinism measure (I8, I9) was recorded\\
RQ4 Governance under machine pace & K4 and K6 blocking gates & D3 measured from objective to approved decision, with gate cost inside the interval & Partially supported. The gates did not eliminate the latency advantage, but I11 and I12 were not captured, so the cost of the gates is not isolated\\\bottomrule
\end{tabularx}
\end{table*}

\subsection{Design Principles Operationalized}
Four design principles are operationalized. First, reasoning and execution should be coupled rather than run as separate phases; C4-L0 realizes this by sequencing both within one task graph. Second, transitions should carry declared preconditions rather than merely structured payloads; K1--K8 make the boundary itself checkable, and what transfers to another domain is the typed boundary rather than the logistics model. Third, systems should correct errors during execution and record learning updates explicitly; C5-S6 and K8 require a cycle identifier and an enumeration of changed parameters rather than silent adaptation. Fourth, authoritative grounding should be consulted before publication rather than used only for post-hoc checking; C8 and K5 force unmatched claims onto an explicit review list.

These principles are paired with measurements rather than asserted as causal outcomes. Analyst time and systems touched measure reconciliation burden; spatial-reference, unsupported-claim, and detection-stage metrics test grounding; provenance completeness and re-query determinism test reproducibility; intervention rate and gate dwell time expose the operational cost of governance. This pairing is important because the deployment observations are aggregate outcomes and cannot isolate individual mechanisms.

\subsection{Limits and Boundary Conditions}
The architecture externalizes some expert judgment as validation conditions but does not capture tacit competence. Examples include recognizing when a nominally valid route is operationally unrealistic, deciding which anomaly warrants escalation in a specific commercial relationship, or weighing a marginal reallocation against customer commitments. HITL gates preserve that expertise in the operating loop rather than claiming to encode it. Systematically eliciting and representing organizational decision knowledge remains a prerequisite for safely raising the autonomy level.

The framework also assumes effective integration with operational GIS execution tools. Robust error detection, interoperability, and scalability under production loads are engineering problems that the architecture organizes but does not itself solve. Deployment-scale utilization and commercial cost figures are confidential. Principal resource drivers are telemetry/index processing, spatio-temporal analysis, model inference, and versioned provenance storage; no general cost or throughput envelope is claimed. A second boundary condition is authoritative-foundation coverage. If C8 cannot resolve enough claims at the operator-accepted precision floor, the unverified-claim list grows and human review becomes the bottleneck; in that regime the architecture relocates coordination burden instead of reducing it. Finally, the evidence is from one operating supply-chain context and does not establish cross-domain generality.

\section{Future Research}
The immediate priority is attributable evaluation. A grounding ablation disables K5 and Claim Validation while retaining contracts and gates, measuring spatial-reference errors, unsupported claims, review load, and failure-detection stage. A governance ablation converts K4/K6 from blocking to advisory, measuring claim errors, intervention rate, gate dwell, and decision latency. A contract ablation replaces typed contracts with ad hoc handoffs while retaining grounding and gates, measuring analyst time, systems touched, and contract failures.

Beyond ablation, the workflow should be evaluated against GeoAgentBench \cite{yu2026} and GeoBenchX \cite{krechetova2025}; repeated under different data-certification regimes; extended with checkable representations of tacit decision knowledge; and studied with logistics planners to test whether gate and unverified-claim interfaces remain usable at operational volume. Broader research should also address quantitative spatial reasoning over multi-step dependencies \cite{dorobantu2026} and additional modalities such as remote sensing and temporal data \cite{talemi2026,hashemi2026}.

\section{Conclusion}
ANASSA is a reference architecture for autonomous geospatial intelligence across distributed systems. Its operative properties are typed agent boundaries with declared preconditions, authoritative validation of spatial claims, and blocking human accountability at consequential decision points. In one operating supply-chain deployment, the architecture was associated with 25\% improved data quality, 50\% lower spatial workflow runtime, and 60\% lower decision latency against a manual baseline. These are uncontrolled observations; absolute deployment-scale quantities remain confidential, and no mechanism-level causal attribution is claimed. The specified instrumentation and ablations define the path from deployment evidence to controlled evaluation. ANASSA does not propose replacing human spatial expertise; it automates coordination labor while retaining accountable human decisions where consequences are highest.

\begin{acks}
During manuscript preparation, the authors used AI tools (Claude by Anthropic and Microsoft Copilot) to assist in refining text and figure ideation. All content was reviewed and finalized by the authors.
\end{acks}
\bibliographystyle{ACM-Reference-Format}
\bibliography{references}

@misc{bao2026,
  author = {Bao, R. and Yang, C. and Yu, D. and Tang, Z. and Mai, G. and Zhao, L.},
  title = {Spatial-Agent: Agentic Geo-Spatial Reasoning with Scientific Core Concepts},
  year = {2026},
  eprint = {2601.16965},
  archivePrefix = {arXiv}
}

@inproceedings{bhattaram2025,
  author = {Bhattaram, A. and Chung, J. and Chung, S. and Gupta, R. and Ramamoorthy, J. and Gullapalli, K. and Marculescu, D. and Stamoulis, D.},
  title = {GeoFlow: Agentic Workflow Automation for Geospatial Tasks},
  booktitle = {Proceedings of the 33rd ACM International Conference on Advances in Geographic Information Systems},
  pages = {1150--1153},
  year = {2025},
  publisher = {Association for Computing Machinery},
  address = {New York, NY, USA},
  doi = {10.1145/3748636.3763217}
}

@misc{cognilytica2019,
  author = {{Cognilytica}},
  title = {{CPMAI}: Cognitive Project Management for {AI} Methodology},
  year = {2019},
  howpublished = {Cognilytica Research}
}

@article{dorobantu2026,
  author = {Dorobantu, G. I. and Badea, A. C.},
  title = {Geospatial Reasoning and Awareness in Large Language Models: A Systematic Review},
  journal = {Artificial Intelligence Review},
  volume = {59},
  number = {4},
  pages = {111},
  year = {2026},
  doi = {10.1007/s10462-026-11512-x}
}

@misc{hashemi2026,
  author = {Hashemi, Mohammad and Amiri, Hossein and Z{\"u}fle, Andreas},
  title = {A Comprehensive Survey of Agentic {AI} for Geospatial Data},
  year = {2026},
  note = {Preprints.org, Version 2},
  doi = {10.20944/preprints202601.2236.v2}
}

@inproceedings{hu2025,
  author = {Hu, J. and Sun, L. and Liu, X.},
  title = {GeoAgent: An Agentic {AI} Framework for Spatial Query Understanding and Interactive Geospatial Intelligence},
  booktitle = {Proceedings of the 1st ACM SIGSPATIAL International Workshop on Generative and Agentic AI for Multi-Modality Space-Time Intelligence},
  pages = {54--60},
  year = {2025},
  publisher = {Association for Computing Machinery},
  address = {New York, NY, USA},
  doi = {10.1145/3764915.3770719}
}

@article{kaza2026,
  author = {Khosravi Kazazi, Ali and Li, Zhenlong},
  title = {From Tool to Collaborator: Rethinking {GIScience} as {GIS} Becomes Agentic and Autonomous},
  journal = {International Journal of Digital Earth},
  volume = {19},
  number = {1},
  pages = {2687228},
  year = {2026},
  doi = {10.1080/17538947.2026.2687228}
}

@inproceedings{krechetova2025,
  author = {Krechetova, Varvara and Kochedykov, Denis},
  title = {GeoBenchX: Benchmarking {LLMs} in Agent Solving Multistep Geospatial Tasks},
  booktitle = {Proceedings of the 1st ACM SIGSPATIAL International Workshop on Generative and Agentic AI for Multi-Modality Space-Time Intelligence},
  pages = {27--35},
  year = {2025},
  publisher = {Association for Computing Machinery},
  address = {New York, NY, USA},
  doi = {10.1145/3764915.3770721}
}

@article{liang2026,
  author = {Liang, C. and Cui, Y. and Shi, R. and Zha, G. and Yin, X. and Xiao, M. and Xu, D. and Duan, X. and Huang, B.},
  title = {GeoAgentic-RAG: A Multi-Agent Framework for Autonomous Geospatial Reasoning and Visual Insight Generation with {LLM}},
  journal = {International Journal of Applied Earth Observation and Geoinformation},
  volume = {147},
  pages = {105195},
  year = {2026},
  doi = {10.1016/j.jag.2026.105195}
}

@article{mansourian2026,
  author = {Mansourian, A. and Oucheikh, R.},
  title = {Bridging Natural Language and {GIS}: A Multi-Agent Framework for {LLM}-Driven Autonomous Geospatial Analysis},
  journal = {International Journal of Digital Earth},
  volume = {19},
  number = {1},
  pages = {2633849},
  year = {2026},
  doi = {10.1080/17538947.2026.2633849}
}

@article{mercier2017,
  author = {Mercier, S. and Villeneuve, S. and Mondor, M. and Uysal, I.},
  title = {Time--Temperature Management along the Food Cold Chain: A Review of Recent Developments},
  journal = {Comprehensive Reviews in Food Science and Food Safety},
  volume = {16},
  number = {4},
  pages = {647--667},
  year = {2017},
  doi = {10.1111/1541-4337.12269}
}

@article{ndraha2018,
  author = {Ndraha, N. and Hsiao, H.-I. and Vlajic, J. and Yang, M.-F. and Lin, H.-T. V.},
  title = {Time-Temperature Abuse in the Food Cold Chain: Review of Issues, Challenges, and Recommendations},
  journal = {Food Control},
  volume = {89},
  pages = {12--21},
  year = {2018},
  doi = {10.1016/j.foodcont.2018.01.027}
}

@phdthesis{papantoniou2021geoblockchain,
  author = {Papantoniou, C.},
  title = {GeoBlockchain: The Analysis, Design, and Evaluation of a Spatially Enabled Blockchain},
  school = {Claremont Graduate University},
  year = {2021},
  note = {CGU Theses and Dissertations 284}
}

@inproceedings{papantoniou2021enterprise,
  author = {Papantoniou, C. and Hilton, B.},
  title = {Enterprise Solutions Criteria in the Age of {GeoBlockchain}: Land Ownership and Supply Chain},
  booktitle = {Proceedings of the 54th Hawaii International Conference on System Sciences},
  pages = {5307--5316},
  year = {2021},
  publisher = {University of Hawai'i at Manoa},
  address = {Honolulu, HI, USA},
  doi = {10.24251/HICSS.2021.647}
}

@misc{talemi2026,
  author = {Talemi, N. A. and Boone, J. and Afghah, F.},
  title = {Agentic {AI} in Remote Sensing: Foundations, Taxonomy, and Emerging Systems},
  year = {2026},
  eprint = {2601.01891},
  archivePrefix = {arXiv}
}

@article{wang2024,
  author = {Wang, L. and Ma, C. and Feng, X. and Zhang, Z. and Yang, H. and Zhang, J. and Chen, Z. and Tang, J. and Chen, X. and Lin, Y. and Zhao, W. and Wei, Z. and Wen, J.},
  title = {A Survey on Large Language Model Based Autonomous Agents},
  journal = {Frontiers of Computer Science},
  volume = {18},
  number = {6},
  pages = {186345},
  year = {2024},
  doi = {10.1007/s11704-024-40231-1}
}

@inproceedings{wu2025,
  author = {Wu, C.-C. and Huang, S.-B. and Song, Y.-L. and Lin, P.-H. and Lin, M. C.-L. and Lin, Y.-T.},
  title = {Democratizing Multi-Granularity Spatio-Temporal Intelligence with Multi-Agent Systems},
  booktitle = {Proceedings of the 1st ACM SIGSPATIAL International Workshop on Generative and Agentic AI for Multi-Modality Space-Time Intelligence},
  pages = {22--26},
  year = {2025},
  publisher = {Association for Computing Machinery},
  address = {New York, NY, USA},
  doi = {10.1145/3764915.3770718}
}

@misc{xi2023,
  author = {Xi, Z. and Chen, W. and Guo, X. and He, W. and Ding, Y. and Hong, B. and Zhang, M. and Wang, J. and Jin, S. and Zhou, E. and Zheng, R. and Fan, X. and Wang, X. and Xiong, L. and Zhou, Y. and Wang, W. and Jiang, C. and Zou, Y. and Liu, X. and Gui, T.},
  title = {The Rise and Potential of Large Language Model Based Agents: A Survey},
  year = {2023},
  eprint = {2309.07864},
  archivePrefix = {arXiv},
  doi = {10.48550/arXiv.2309.07864}
}

@misc{yu2026,
  author = {Yu, Bo and Yang, Cheng and Hou, Dongyang and Liu, Chengfu and Liu, Jiayao and Wang, Chi and Zhang, Zhiming and Li, Haifeng and Yang, Wentao},
  title = {GeoAgentBench: A Dynamic Execution Benchmark for Tool-Augmented Agents in Spatial Analysis},
  year = {2026},
  eprint = {2604.13888},
  archivePrefix = {arXiv},
  primaryClass = {cs.AI},
  doi = {10.48550/arXiv.2604.13888}
}
\end{document}